\documentclass[english]{article}
\usepackage[T1]{fontenc}
\usepackage[latin9]{inputenc}
\usepackage{color}
\usepackage{graphicx}
\usepackage{setspace}
\usepackage[numbers]{natbib}

\makeatletter

\providecommand{\tabularnewline}{\\}

\newcommand{\lyxaddress}[1]{
\par {\raggedright #1
\vspace{1.4em}
\noindent\par}
}

\usepackage{babel}

\usepackage{babel}

\usepackage{babel}

\usepackage{babel}

\makeatother

\usepackage{babel}
\begin{document}

\title{Image Splicing Localization Using A Multi-Task Fully Convolutional
Network (MFCN)}

\author{Ronald Salloum, Yuzhuo Ren and C.-C. Jay Kuo}
\maketitle
\begin{singlespace}

\lyxaddress{\begin{center}
Ming Hsieh Department of Electrical Engineering, University of Southern
California, Los Angeles, CA
\par\end{center}}
\end{singlespace}
\begin{abstract}
In this work, we propose a technique that utilizes a fully convolutional
network (FCN) to localize image splicing attacks. We first evaluated
a single-task FCN (SFCN) trained only on the surface label. Although
the SFCN is shown to provide superior performance over existing methods,
it still provides a coarse localization output in certain cases. Therefore,
we propose the use of a multi-task FCN (MFCN) that utilizes two output
branches for multi-task learning. One branch is used to learn the
surface label, while the other branch is used to learn the edge or
boundary of the spliced region. We trained the networks using the
CASIA v2.0 dataset, and tested the trained models on the CASIA v1.0,
Columbia Uncompressed, Carvalho, and the DARPA/NIST Nimble Challenge
2016 SCI datasets. Experiments show that the SFCN and MFCN outperform
existing splicing localization algorithms, and that the MFCN can achieve
finer localization than the SFCN.
\end{abstract}

\section{Introduction}

\label{sec:intro}

With the advent of Web 2.0 and ubiquitous adoption of low-cost and
high-resolution digital cameras, users upload and share images on
a daily basis. This trend of public image distribution and access
to user-friendly editing software such as Photoshop and GIMP has made
image forgery a serious issue. Splicing is one of the most common
types of image forgery. It manipulates images by copying a region
from one image (i.e., the donor image) and pasting it onto another
image (i.e., the host image). Forgers often use splicing to give a
false impression that there is an additional object present in the
image, or to remove an object from the image.

\begin{figure}
\centering{} \includegraphics[width=0.58\textwidth]{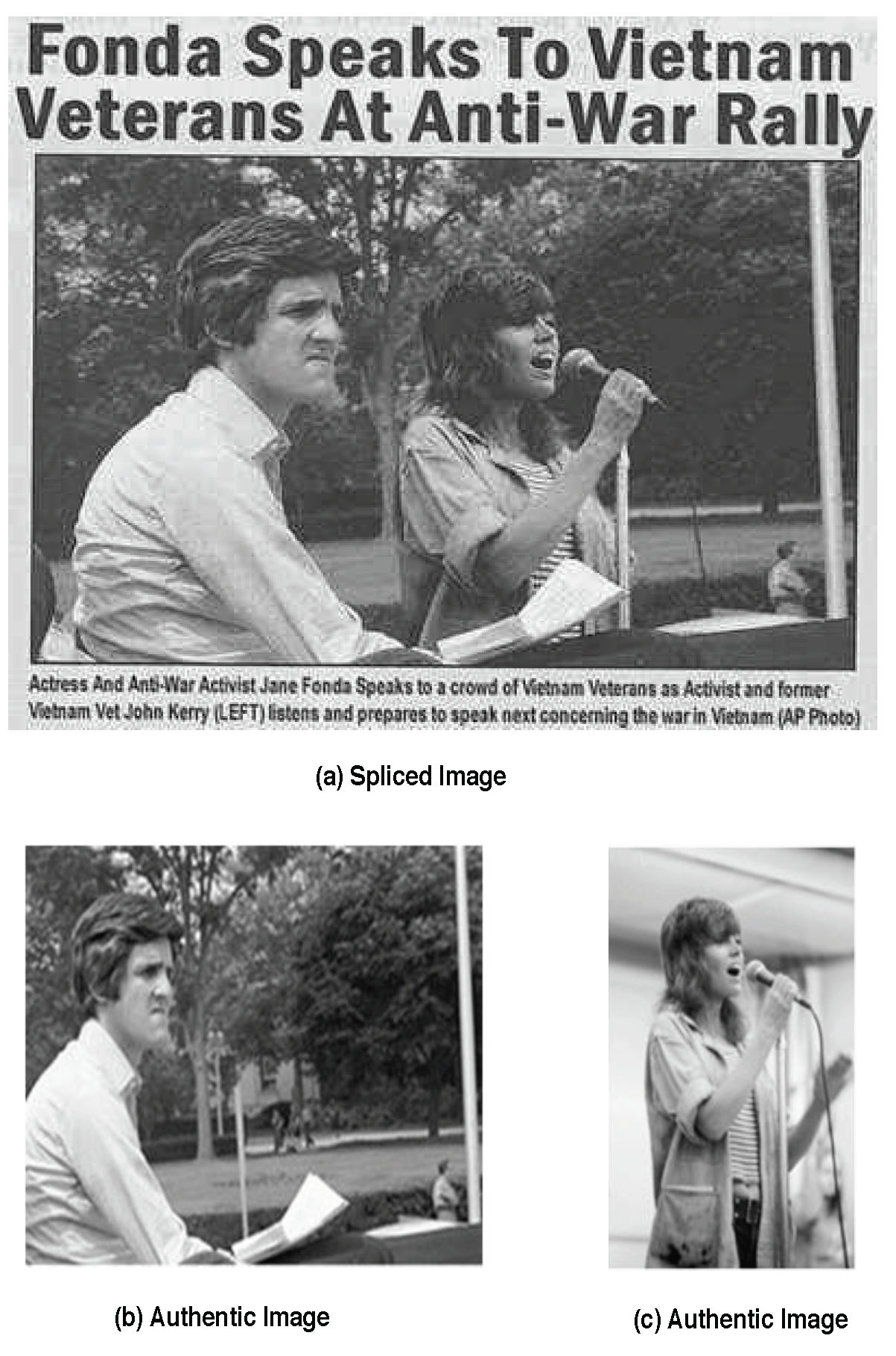}
\caption{An image splicing example \citep{shi2007natural}: (a) the spliced
image showing John Kerry and Jane Fonda together at an anti-Vietnam
war rally, (b) an authentic image of Kerry, and (c) an authentic image
of Fonda.}
\label{fig:Kerry} 
\end{figure}

\begin{figure}
\centering{}\centering{}\includegraphics[clip,width=0.8\textwidth]{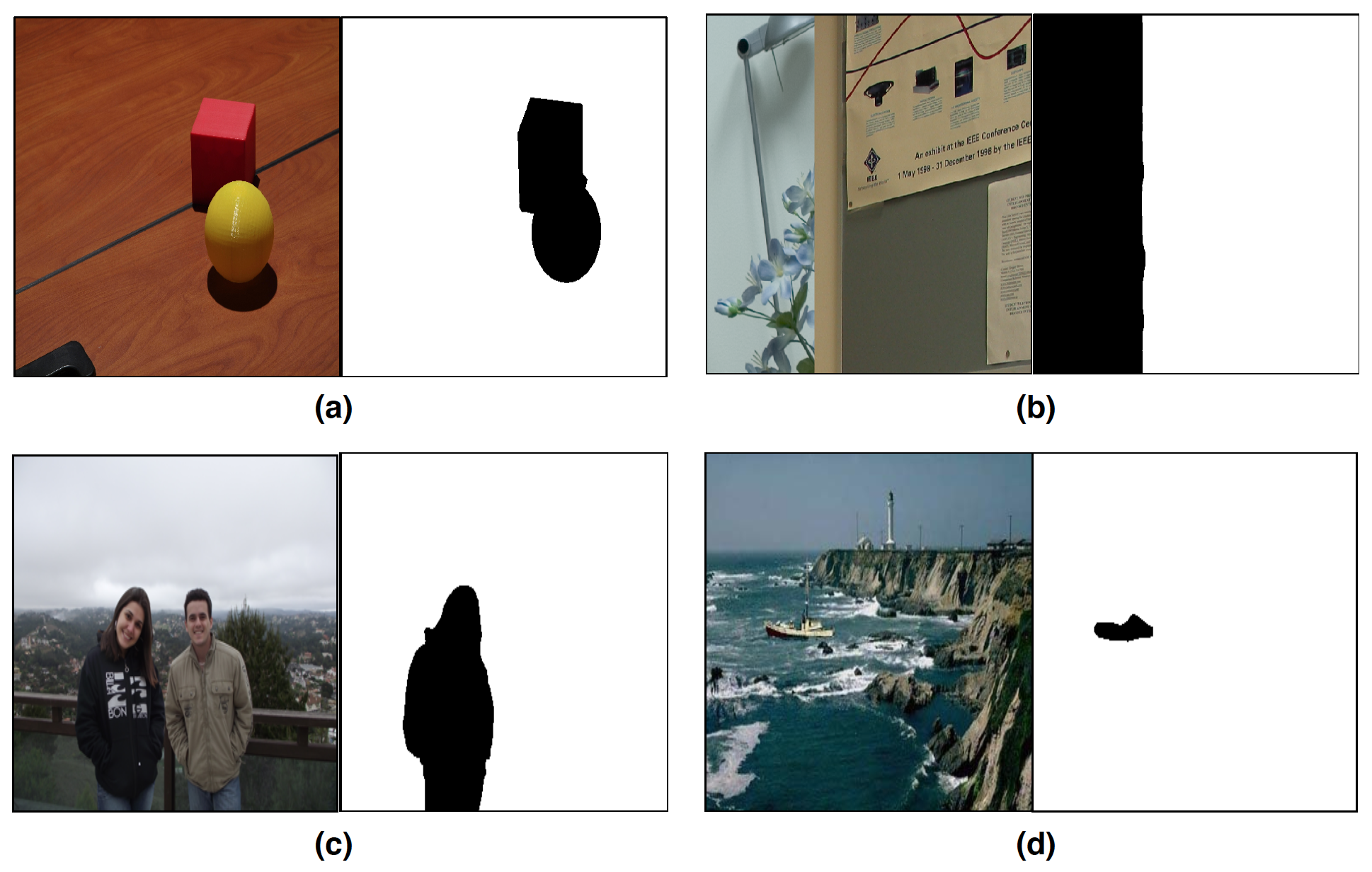}
\caption{Image splicing examples and corresponding ground truth masks obtained
from four different datasets: (a) the Nimble Science (SCI) dataset,
(b) the Columbia Uncompressed dataset, (c) the Carvalho dataset, and
(d) the CASIA v1.0 dataset. For the ground truth mask, pixels that
were manipulated are represented by a value of 0 (the black region)
and pixels that were not manipulated are represented by a value of
255 (the white region).}
\label{Splicing-Examples} 
\end{figure}

Image splicing can be potentially used in generating false propaganda
for political purposes. For example, during the 2004 US Presidential
election campaign, an image that showed John Kerry and Jane Fonda
speaking together at an anti-Vietnam war protest was released and
circulated. It was discovered later that this was a spliced image,
and was created for political purposes. Fig. \ref{fig:Kerry} shows
the spliced image, along with the two original authentic images that
were used to create the spliced image \citep{shi2007natural}. Some
additional splicing examples obtained from four datasets are shown
in Fig. \ref{Splicing-Examples}.

%%%%%%%%%%%%%%%%%%%%%%%%%%%%%%%%%%

%%%%%%%%%%%%%%%%%%%%%%%%%%%%%%%%%%

\textcolor{black}{Many of the current splicing detection algorithms
only deduce whether a given image has been spliced and do not attempt
to localize the spliced area. Relatively few algorithms attempt to
tackle the splicing localization problem, which refers to the problem
of determining which pixels in an image have been manipulated as a
result of a splicing operation. A brief review of existing splicing
localization techniques will be given in Sec. \ref{sec:review}.}

In this paper, we present an effective solution to the splicing localization
problem based on a fully convolutional network (FCN). The base network
architecture is the FCN VGG-16 architecture with skip connections,
but we incorporate several modifications, including batch normalization
layers and class weighting. We first evaluated a single-task FCN (SFCN)
trained only on the surface label or ground truth mask, which classifies
each pixel in a spliced image as spliced or authentic. Although the
SFCN is shown to provide superior performance over existing techniques,
it still provides a coarse localization output in certain cases. Thus,
we next propose the use of a multi-task FCN (MFCN) that utilizes two
output branches for multi-task learning. One branch is used to learn
the surface label, while the other branch is used to learn the edge
or boundary of the spliced region. It is shown that by simultaneously
training on the surface and edge labels, we can achieve finer localization
of the spliced region, as compared to the SFCN. Once the MFCN was
trained, we evaluated two different inference approaches. The first
approach utilizes only the surface output probability map in the inference
step. The second approach, which is referred to as the edge-enhanced
MFCN, utilizes both the surface and edge output probability maps to
achieve finer localization. We trained the SFCN and MFCN using the
CASIA v2.0 dataset and tested the trained networks on the CASIA v1.0,
Columbia Uncompressed \citep{hsu2006detecting}, Carvalho \citep{de2013exposing},
and the DARPA/NIST Nimble Challenge 2016 Science (SCI) datasets \footnote{https://www.nist.gov/itl/iad/mig/nimble-challenge}.
Experiments show that the SFCN and MFCN outperform existing splicing
localization algorithms, with the edge-enhanced MFCN achieving the
best performance. Furthermore, we show that after applying various
post-processing techniques such as JPEG compression, blurring, and
added noise to the spliced images, the SFCN and MFCN methods still
outperform the existing methods.

%%%%%%%%%%%%%%%%%%%%%%%%%%%%%%%%%%

%%%%%%%%%%%%%%%%%%%%%%%%%%%%%%%%%%

The rest of the paper is organized as follows. Related work is reviewed
in Sec. \ref{sec:review}. The proposed methods are described in Sec.
\ref{sec:MFCN}. The performance evaluation metrics are discussed
in Sec. \ref{sec:metrics}. Experimental results are presented in
Sec. \ref{sec:experiments}. Finally, concluding remarks are given
in Sec. \ref{sec:conclusion}.

\section{Related Work}

\label{sec:review}

An extensive review of splicing localization techniques and a comparison
of their performance can be found in a recent paper by Zampoglou {\em
et al.} \citep{zampoglou2016large}. We provide a brief summary in
this section. Existing splicing localization algorithms can be roughly
divided into three classes based on the pattern or trace types used
to separate the spliced region from the rest of the image. They exploit
the following traces (or features): 1) noise patterns, 2) Color Filter
Array (CFA) interpolation patterns, and 3) JPEG-related traces.

The first class of splicing localization algorithms exploits noise
patterns under the assumption that different images have different
noise patterns as a result of a combination of different camera makes/models,
the capture parameters of each image, and post-processing techniques
\citep{lyu2014exposing,mahdian2009using,cozzolino2015splicebuster,chen2008determining,chierchia2014bayesian,li2012color}.
Since the spliced region originated from a different image (i.e.,
the donor image) then the host image, the spliced region may have
a noise pattern that is different than the noise pattern in the remaining
region of the host image. Thus, the noise pattern can potentially
be used to identify the spliced region.

The second class of algorithms exploits CFA interpolation patterns
\citep{dirik2009image,ferrara2012image}. Most digital cameras acquire
images using a single image sensor overlaid with a CFA that produces
one value per pixel. CFA interpolation (also called demosaicing) is
a process to reconstruct the full color image by transforming the
captured output into three channels (RGB). Splicing can disrupt the
CFA interpolation patterns in multiple ways. For example, different
cameras may use different CFA interpolation algorithms so that combining
two different images may cause discontinuities. Also, spliced regions
are often rescaled, which can also disrupt the CFA interpolation patterns.
These artifacts can be used when attempting to localize a spliced
region.

The third class of algorithms exploits the traces left by JPEG compression.
Most of these methods use features from one of two subgroups: JPEG
quantization artifacts and JPEG compression grid discontinuities \citep{lin2009fast,bianchi2012detection,bianchi2012image,farid2009exposing,li2009passive,ye2007detecting,luo2007novel,amerini2014splicing,bianchi2011improved}.
In JPEG quantization-based methods, it is assumed that the original
image underwent consecutive JPEG compressions, while the spliced portion
may have lost its initial JPEG compression characteristics due to
smoothing or resampling of the spliced portion. These incongruous
features can help localize a spliced region. In JPEG grid-based methods,
one may detect spliced regions due to misalignment of the 8x8 block
grids used in compression. Two other approaches that exploit JPEG
compression traces are JPEG Ghosts \citep{farid2009exposing} and
Error Level Analysis \citep{krawetz2007pictures,wang2010tampered,patel2015improvement}. 

\textcolor{black}{In this work, we propose a neural-network-based
solution, which does not require any feature extraction. Instead,
the relevant features are automatically learned during the network
training phase. The introduction of neural-network-based solutions
has changed the landscape of the computer vision field significantly
in the last five years. In the following section, we present such
a solution for the image splicing localization problem. Of the existing
techniques in the published literature on image forensics, relatively
few have utilized deep learning or neural-network-based techniques
\citep{rao2016deep,zhang2016image,bayar2016deep,cozzolino2016single}.
These techniques have either (1) not focused specifically on splicing
attacks, (2) did not report quantitative results using a pixel-wise
localization metric (e.g., some papers reported only detection-based
metrics) or (3) did not use a public splicing dataset when reporting
pixel-wise localization results.}

\section{Proposed Methods}

\label{sec:MFCN}

\subsection{Brief Review of Fully Convolutional Networks (FCNs)}

In this work, we propose an image splicing localization technique
based on a convolutional neural network (CNN). CNNs have been shown
to provide promising results in image-based detection tasks such as
object detection. Fully convolutional networks (FCNs) \citep{long2015fully}
are a special type of convolutional neural network with only convolutional
layers (i.e., by converting all fully connected layers to convolutional
ones). In \citep{long2015fully}, the authors adapted common classification
networks into fully convolutional ones for the task of semantic segmentation.
It was shown in \citep{long2015fully} that FCNs can efficiently learn
to make dense predictions for per-pixel tasks such as semantic segmentation.
Three classification architectures that were converted to fully convolutional
form are AlexNet \citep{krizhevsky2012imagenet}, GoogLeNet \citep{szegedy2015going},
and VGG-16 \citep{simonyan2014very}. In \citep{long2015fully}, the
authors found that FCN VGG-16 performed the best among the three. 

\subsection{Single-task Fully Convolutional Network (SFCN)}

\label{subsec:SFCN}

\textcolor{black}{In this work, we first adapted a single-task fully
convolutional network (SFCN) based on the FCN VGG-16 architecture
to the spicing localization problem. The SFCN was trained on the surface
label or ground truth mask, which is a per-pixel, binary mask that
classifies each pixel in an image as spliced or authentic. We utilized
the skip architecture proposed in \citep{long2015fully}. In \citep{long2015fully},
the authors extended the FCN VGG-16 to a three-stream network with
eight pixel prediction stride (as opposed to the 32 pixel stride in
the original network without skip connections), and found that this
skip architecture yielded superior performance over the original architecture.
In addition, we incorporated several modifications, including batch
normalization and class weighting. We utilized batch normalization
to eliminate the bias and normalize the inputs at each layer \citep{ioffe2015batch}.
Class weighting refers to the application of different weights to
the different classes in the loss function. In the context of splicing
localization, class weighting is beneficial since the amount of non-spliced
(authentic) pixels typically outnumbers the amount of spliced pixels
by a significant margin. We apply a larger weight to the spliced pixels
(since there are fewer spliced pixels than non-spliced ones). In particular,
we used median frequency class weighting \citep{eigen2015predicting,badrinarayanan2015segnet}.
Although the ground truth mask is binary, the raw output of the SFCN
is a probability map. That is, each pixel is assigned a value representing
the probability that the pixel is spliced. As described in Section
\ref{sec:metrics}, the performance metrics require a binary mask,
thus the probability map must be thresholded. We refer to the thresholded
probability map as the binary system output mask.}

\subsection{Multi-task Fully Convolutional Network (MFCN)}

\label{subsec:MFCN}

\begin{figure}
\centering{}\textcolor{black}{\centering{} \includegraphics[width=1\textwidth]{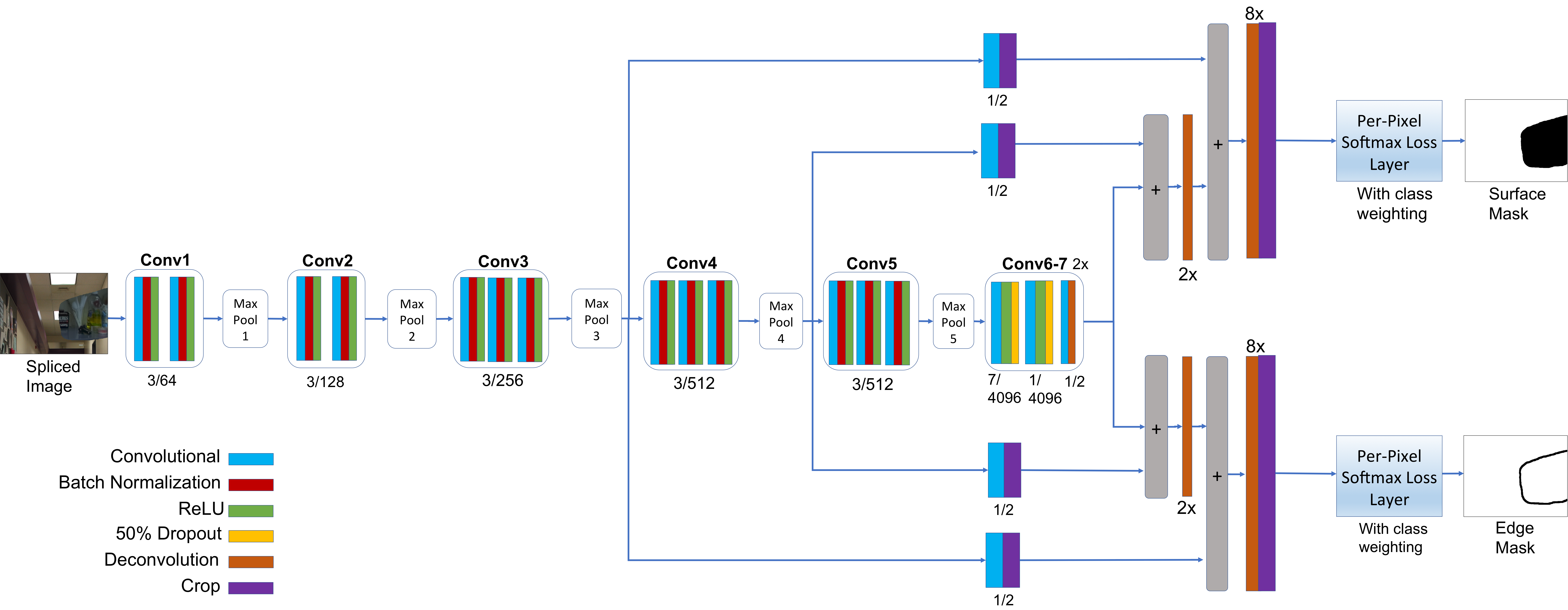}
\caption{The MFCN Architecture for image splicing localization. Numbers in
the form x/y refer to the kernel size and number of filters of the
convolutional layer (colored blue), respectively. For example, the
Conv1 block consists of two convolutional layers, each with a kernel
size of 3 and 64 filters (note that after each convolutional layer
is a batch normalization layer and a ReLU layer). Numbers in the form
of 2x and 8x refer to an upsampling factor of 2 and 8 for the deconvolution
layers, respectively. Also, please note the inclusion of skip connections
at the third and fourth max pooling layers. The grey-colored layers
represent element-wise addition. The max pooling layers have a kernel
size of 2 and a stride of 2.}
\label{MFCN Architecture} }
\end{figure}

\textcolor{black}{Although the single-task network is shown to provide
superior performance over existing splicing localization methods,
it can still provide a coarse localization output in certain cases.
In \citep{ren2016coarse_v1}, a multi-task fully convolutional network
(MFCN) based on the FCN VGG-16 architecture was proposed for indoor
layout estimation. The multi-task architecture utilizes two output
branches instead of one and it offers superior performance for the
indoor layout estimation problem. In our work, we adopt the idea in
\citep{ren2016coarse_v1} of utilizing a multi-task network, but we
incorporate several modifications, including skip connections, batch
normalization, and class weighting (as discussed in Section \ref{subsec:SFCN}).
In contrast to the SFCN, the MFCN utilizes two output branches for
multi-task learning. One branch is used to learn the surface label,
while the other branch is used to learn the edge or boundary of the
spliced region. The architecture of the MFCN used in our paper is
shown in Fig. \ref{MFCN Architecture}. In addition to the surface
labels, the boundaries between inserted regions and their host background
can be an important indicator of a manipulated area. This is what
motivated us to use a multi-task learning network. The weights or
parameters of the network are influenced by both the surface and edge
labels during the training process. By simultaneously training on
the surface and edge labels, we are able to obtain a finer localization
of the spliced region, as compared to training only on the surface
labels. Once the network was fully trained, we evaluated two different
binary output mask generation approaches. In the first approach, we
extract the surface output probability map, and then threshold it
to yield the binary system output mask. In this approach, the edge
output probability map is not utilized in the inference step. Please
note that the edge label still influenced the weights of the network
during the training process. }

\subsection{Edge-enhanced MFCN Inference}

\begin{figure}
\centering{}\textcolor{black}{\centering{} \includegraphics[width=0.95\textwidth]{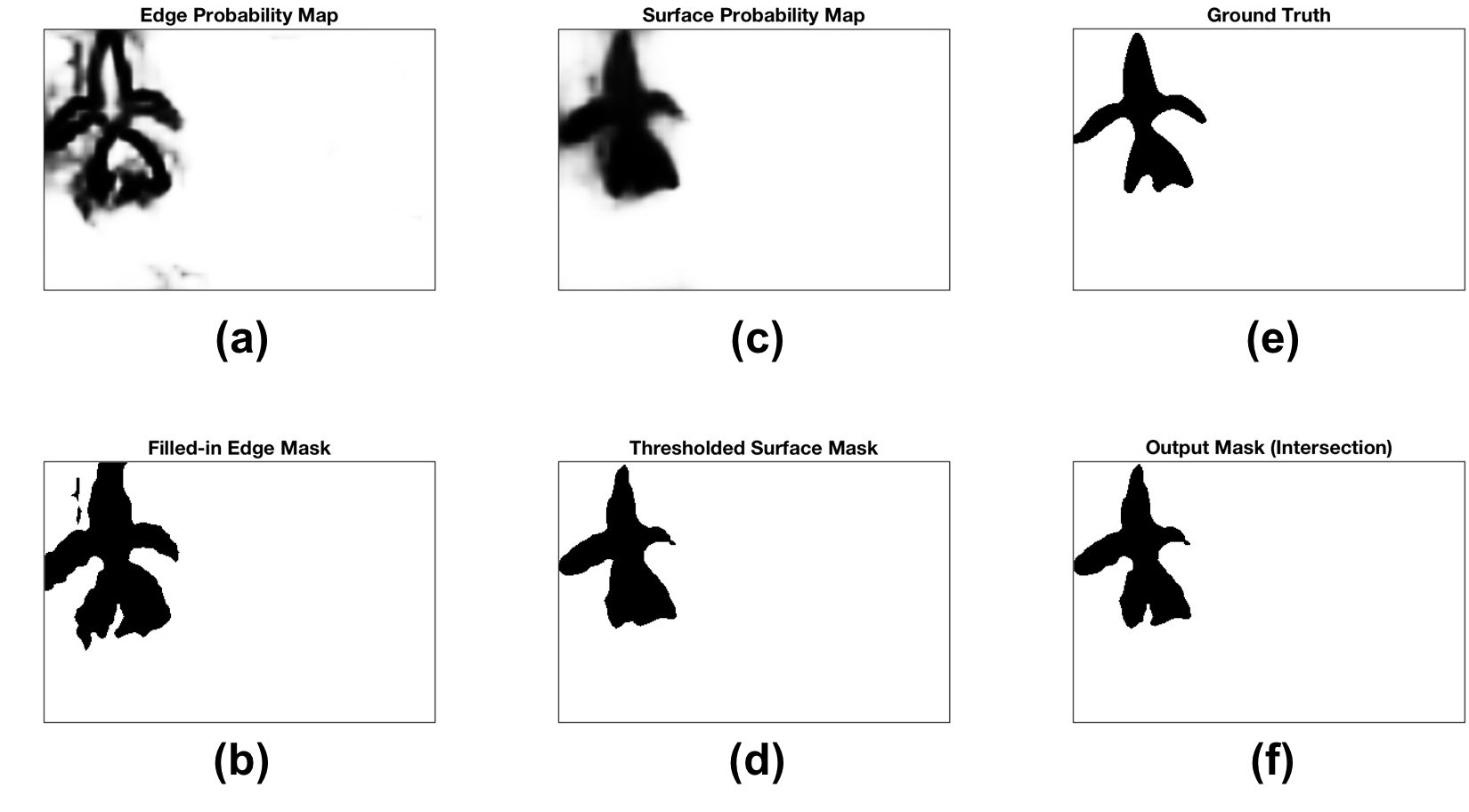}
\caption{\textcolor{black}{Illustration of MFCN inference with edge enhancement:
(a) Edge probability map, (b) Hole-filled, thresholded edge mask,
(c) Surface probability map, (d) Thresholded surface mask, (e) Ground
truth mask, and (f) Final system output mask.}}
\label{edge_enhancement} }
\end{figure}

\textcolor{black}{The second inference strategy, which we refer to
as the edge-enhanced MFCN, utilizes both the surface and edge output
probability maps, as described in the following steps: }
\begin{enumerate}
\item \textcolor{black}{We threshold the surface probability map with a
given threshold. }
\item \textcolor{black}{We threshold the edge probability map with a given
threshold. }
\item \textcolor{black}{Next, we apply hole-filling to the output of step
(2), yielding the hole-filled, thresholded edge mask.}
\item \textcolor{black}{Finally, we generate the binary system output mask
by computing the intersection of the output of step (1) and output
of step (3). }
\end{enumerate}
\textcolor{black}{It is shown in this paper that by utilizing both
the edge and surface probability maps in the inference step, we obtain
finer localization of the spliced region. An example illustrating
inference with edge-enhancement is shown in Figure \ref{edge_enhancement}.
It can be seen that utilizing both the edge and surface probability
maps leads to a finer localization of the spliced region.}

%%%%%%%%%%%%%%%%%%%%%%%%%%%%%%%%%%

%%%%%%%%%%%%%%%%%%%%%%%%%%%%%%%%%%

\subsection{Training and Testing Procedure}

For the MFCN, the total loss function, $L_{t}$, is the sum of the
loss corresponding to the surface label and the loss corresponding
to the edge label, denoted by $L_{s}$ and $L_{e}$, respectively.
Thus, we have 
\[
L_{t}=L_{s}+L_{e},
\]
where $L_{s}$ and $L_{e}$ are per-pixel softmax loss functions.
In addition, we apply median-frequency class weighting to the surface
and edge loss functions, as described in Sections \ref{subsec:SFCN}
and \ref{subsec:MFCN}. For the SFCN, the total loss function is equal
to the surface loss function $L_{s}$.

The training of the networks was performed in Caffe \citep{jia2014caffe}
using the stochastic gradient descent (SGD) algorithm with a fixed
learning rate of 0.0001, a momentum of 0.9, and a weight decay of
0.0005. We initialize the weights or parameters of the SFCN and MFCN
by the weights of a VGG-16 model pre-trained on the ImageNet dataset,
which contains 1.2 million images for the task of object recognition
and image classification. \citep{simonyan2014very,russakovsky2015imagenet}. 

We trained the SFCN and MFCN using the CASIA v2.0 dataset, and then
evaluated the trained models on the CASIA v1.0, Columbia Uncompressed,
Carvalho, and Nimble Challenge 2016 Science (SCI) datasets. The numbers
of training and testing images are given in Table \ref{Training_Testing_Numbers}.
Ground truth masks are provided for the Columbia Uncompressed, Carvalho,
and Nimble Challenge 2016 datasets. For the CASIA v1.0 and CASIA v2.0
datasets, we generated the ground truth masks using the provided reference
information. In particular, for a given spliced image in the CASIA
datasets, the corresponding donor and host images are provided, and
we used this information to generate the ground truth mask.

%%%%%%%%%%%%%%%%%%%%%%%%%%%%%%%%%%
\begin{table}
\caption{Training and Testing Images}

\begin{centering}
\begin{tabular}{|c|c|c|}
\hline 
Dataset & Type & Number of Images\tabularnewline
\hline 
\hline 
CASIA v2.0 & Training & 5123\tabularnewline
\hline 
CASIA v1.0 & Testing & 921\tabularnewline
\hline 
Columbia & Testing & 180\tabularnewline
\hline 
Carvalho & Testing & 100\tabularnewline
\hline 
Nimble 2016 SCI & Testing & 160\tabularnewline
\hline 
\end{tabular}
\par\end{centering}
\centering{}\label{Training_Testing_Numbers}
\end{table}

%%%%%%%%%%%%%%%%%%%%%%%%%%%%%%%%%%

\section{Performance Evaluation Metrics}

\label{sec:metrics}

\textcolor{black}{Once the MFCN or SFCN is trained, we use the trained
model to evaluate other images not in the training set. We evaluated
the performance of the proposed and existing methods using the $F_{1}$
and Matthews Correlation Coefficient ($MCC$) metrics, which are per-pixel
localization metrics. }

\textcolor{black}{Both the $F_{1}$ metric and $MCC$ metric require
as input a binary mask. We converted each output map to a binary mask
based on a threshold. For each output map, we varied the threshold
and picked the optimal threshold (this is done for each method). This
technique of varying the threshold was also utilized by Zampoglou
et. al. in \citep{zampoglou2016large}. We then computed the $F_{1}$
and $MCC$ metrics by comparing the binary system output mask to the
corresponding ground truth mask. For a given spliced image, the $F_{1}$
metric is defined as
\[
F_{1}(M_{out},M_{gt})=\frac{2TP}{2TP+FN+FP},
\]
where $M_{out}$ represents the binary system output mask, $M_{gt}$
represents the ground truth mask, $TP$ represents the number of pixels
classified as true positive, $FN$ represents the number of pixels
classified as false negative, and $FP$ represents the number of pixels
classified as false positive. A true positive means that a spliced
pixel is correctly classified as spliced, a false negative means that
a spliced pixel is incorrectly classified as authentic, and a false
positive means that an authentic pixel is incorrectly classified as
spliced. For a given spliced image, the $MCC$ metric is defined as }

\textcolor{black}{
\[
MCC(M_{out},M_{gt})=\frac{TP*TN-FP*FN}{\sqrt{(TP+FP)(TP+FN)(TN+FP)(TN+FN)}}.
\]
For a given dataset and a given method, we report the average $F_{1}$
and average $MCC$ scores across the dataset.}

\section{Experimental Results}

\label{sec:experiments}

\subsection{Performance Comparison}

We compared the proposed SFCN and MFCN methods with a large number
of existing splicing localization algorithms. Following the acronyms
in \citep{zampoglou2016large}, these algorithms are: ADQ1 \citep{lin2009fast},
ADQ2 \citep{bianchi2011improved}, ADQ3 \citep{amerini2014splicing},
NADQ \citep{bianchi2012image}, BLK \citep{li2009passive}, CFA1 \citep{ferrara2012image},
CFA2 \citep{dirik2009image}, DCT \citep{ye2007detecting}, ELA \citep{krawetz2007pictures},
NOI1 \citep{mahdian2009using}, NOI2 \citep{lyu2014exposing} and
NOI3 \citep{cozzolino2015splicebuster}. The implementation of these
existing algorithms is provided in a publicly available Matlab toolbox
written by Zampoglou et. al. \citep{zampoglou2016large}. As noted
in \citep{zampoglou2016large}, ADQ2, ADQ3, and NADQ require JPEG
images as input because they exploit certain JPEG data directly extracted
from the compressed files. Therefore, these three algorithms could
only be evaluated on the CASIA v1.0 and Nimble 2016 SCI datasets,
which contain images in JPEG format. These three algorithms could
not be evaluated on the Columbia and Carvalho datasets, since they
do not contain images in JPEG format.

\textcolor{black}{For each method, we computed the average $F_{1}$
and $MCC$ scores across each dataset, and the results are shown in
Table \ref{F1_scores} ($F_{1}$ scores) and Table \ref{MCC_scores}
($MCC$ scores). We see that the MFCN and SFCN methods outperform
the benchmarking algorithms on all four datasets, in terms of both
$F_{1}$ and $MCC$ score. Furthermore, we see that the edge-enhanced
MFCN method performs the best among the three proposed methods (the
numbers are highlighted in bold in Tables \ref{F1_scores} and \ref{MCC_scores}
to reflect this).}

\begin{figure}
\caption{\textcolor{black}{System Output Mask Examples of SFCN, MFCN, and edge-enhanced
MFCN on the CASIA v1.0 and Carvalho Datasets. Please note that we
refer to the MFCN without edge-enhanced inference simply as MFCN.
Each row in the figure shows a manipulated or spliced image, the ground
truth mask, the SFCN output, the MFCN output, and the edge-enhanced
MFCN output. The number below each output example is the corresponding
$F_{1}$ score. The first two rows are examples from the CASIA v1.0
dataset, while the other two rows are examples from the Carvalho dataset.
It can be seen from these examples that the edge-enhanced MFCN achieves
finer localization than the SFCN and the MFCN without edge-enhanced
inference.}}

\begin{centering}
\includegraphics[width=0.75\columnwidth]{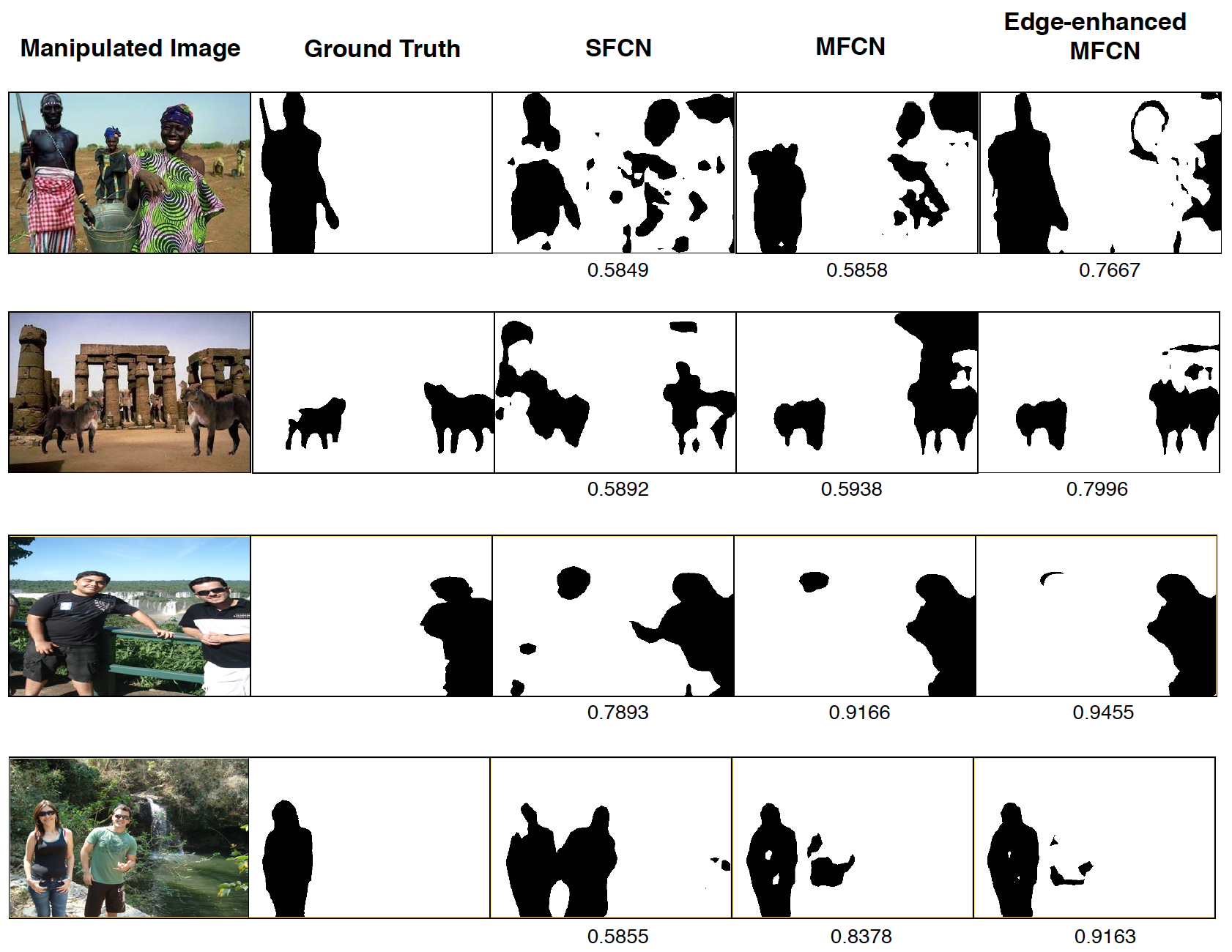}
\par\end{centering}
\label{output_examples_casia1_and_carvalho}
\end{figure}

\textcolor{black}{Since the MFCN and SFCN methods were the top performing
methods, we show system output examples from these methods on the
CASIA v1.0 and Carvalho datasets in Figure \ref{output_examples_casia1_and_carvalho}.
Each row shows a manipulated image, the ground truth mask, and the
binary system output mask for SFCN, MFCN (without edge-enhanced inference),
and the edge-enhanced MFCN. As shown by examples in this figure, the
edge-enhanced MFCN yields finer localization of the spliced region
than the SFCN and the MFCN without edge-enhanced inference. }

\begin{table}
\caption{\textcolor{black}{Average $F_{1}$ Scores of Proposed and Existing
Methods For Different Datasets. For each dataset, we highlight in
bold the top-performing method. As noted in \citep{zampoglou2016large},
ADQ2, ADQ3, and NADQ require JPEG images as input because they exploit
certain JPEG data directly extracted from the compressed files. Therefore,
these three algorithms could only be evaluated on the CASIA v1.0 and
Nimble 2016 SCI datasets, which contain images in JPEG format. For
the Columbia and Carvalho datasets (which do not contain images in
JPEG format), we put ``NA'' in the corresponding entry in the table
to indicate that these three algorithms could not be evaluated on
these two datasets.}}

\noindent \begin{raggedright}
\hspace{-0.15in}%
\begin{tabular}{|c|c|c|c|c|}
\hline 
Method & CASIA v1.0 & Columbia & Nimble 2016 SCI & Carvalho\tabularnewline
\hline 
\hline 
SFCN & 0.4770 & 0.5820 & 0.4220 & 0.4411\tabularnewline
\hline 
MFCN & 0.5182 & 0.6040 & 0.4222 & 0.4678\tabularnewline
\hline 
Edge-enhanced MFCN & \textbf{0.5410} & \textbf{0.6117} & \textbf{0.5707} & \textbf{0.4795}\tabularnewline
\hline 
NOI1 & 0.2633 & 0.5740 & 0.2850 & 0.3430\tabularnewline
\hline 
DCT & 0.3005 & 0.5199 & 0.2756 & 0.3066\tabularnewline
\hline 
CFA2 & 0.2125 & 0.5031 & 0.1587 & 0.3124\tabularnewline
\hline 
NOI3 & 0.1761 & 0.4476 & 0.1635 & 0.2693\tabularnewline
\hline 
BLK & 0.2312 & 0.5234 & 0.3019 & 0.3069\tabularnewline
\hline 
ELA & 0.2136 & 0.4699 & 0.2358 & 0.2756\tabularnewline
\hline 
ADQ1 & 0.2053 & 0.4975 & 0.2202 & 0.2943\tabularnewline
\hline 
CFA1 & 0.2073 & 0.4667 & 0.1743 & 0.2932\tabularnewline
\hline 
NOI2 & 0.2302 & 0.5318 & 0.2320 & 0.3155\tabularnewline
\hline 
ADQ2 & 0.3359 & NA & 0.3433 & NA\tabularnewline
\hline 
ADQ3 & 0.2192 & NA & 0.2622 & NA\tabularnewline
\hline 
NADQ & 0.1763 & NA & 0.2524 & NA\tabularnewline
\hline 
\end{tabular}
\par\end{raggedright}
\label{F1_scores}
\end{table}
 
\begin{table}
\caption{\textcolor{black}{Average $MCC$ Scores of Proposed and Existing Methods
For Various Datasets. For each dataset, we highlight in bold the top-performing
method. As noted in \citep{zampoglou2016large}, ADQ2, ADQ3, and NADQ
require JPEG images as input because they exploit certain JPEG data
directly extracted from the compressed files. Therefore, these three
algorithms could only be evaluated on the CASIA v1.0 and Nimble 2016
SCI datasets, which contain images in JPEG format. For the Columbia
and Carvalho datasets (which do not contain images in JPEG format),
we put ``NA'' in the corresponding entry in the table to indicate
that these three algorithms could not be evaluated on these two datasets.}}

\noindent \begin{raggedright}
\hspace{-0.15in}%
\begin{tabular}{|c|c|c|c|c|}
\hline 
Method & CASIA v1.0 & Columbia & Nimble 2016 SCI & Carvalho\tabularnewline
\hline 
\hline 
SFCN & 0.4531 & 0.4201 & 0.4202 & 0.3676\tabularnewline
\hline 
MFCN & 0.4935 & 0.4645 & 0.4204 & 0.3901\tabularnewline
\hline 
Edge-enhanced MFCN & \textbf{0.5201} & \textbf{0.4792} & \textbf{0.5703} & \textbf{0.4074}\tabularnewline
\hline 
NOI1 & 0.2322 & 0.4112 & 0.2808 & 0.2454\tabularnewline
\hline 
DCT & 0.2516 & 0.3256 & 0.2600 & 0.1892\tabularnewline
\hline 
CFA2 & 0.1615 & 0.3278 & 0.1235 & 0.1976\tabularnewline
\hline 
NOI3 & 0.0891 & 0.2076 & 0.1014 & 0.1080\tabularnewline
\hline 
BLK & 0.1769 & 0.3278 & 0.2657 & 0.1768\tabularnewline
\hline 
ELA & 0.1337 & 0.2317 & 0.1983 & 0.1111\tabularnewline
\hline 
ADQ1 & 0.1262 & 0.2710 & 0.1880 & 0.1493\tabularnewline
\hline 
CFA1 & 0.1521 & 0.2281 & 0.1408 & 0.1614\tabularnewline
\hline 
NOI2 & 0.1715 & 0.3473 & 0.2066 & 0.1919\tabularnewline
\hline 
ADQ2 & 0.3000 & NA & 0.3210 & NA\tabularnewline
\hline 
ADQ3 & 0.1732 & NA & 0.2512 & NA\tabularnewline
\hline 
NADQ & 0.0987 & NA & 0.2310 & NA\tabularnewline
\hline 
\end{tabular}
\par\end{raggedright}
\centering{}\label{MCC_scores}
\end{table}
 
\subsection{Performance on JPEG Compressed Images}

We also compared the performance of the proposed and existing methods
before and after JPEG compression of the spliced images. We used the
Carvalho dataset for this experiment. The images are originally in
PNG format, and we compressed them using two different quality factors:
50 and 70. Table \ref{jpeg_compress} shows the average $F_{1}$ scores
on the original dataset and the JPEG compressed images using the two
different quality factors. We see small performance degradation due
to JPEG compression. However, the performance of the SFCN and MFCN
methods is still better than the performance of existing methods.
Furthermore, we see that the performance of the SFCN and MFCN methods
on the JPEG compressed dataset is better than the performance of existing
methods on the original uncompressed dataset.
\begin{table}
\caption{\textcolor{black}{Average $F_{1}$ Scores of Proposed and Existing
Methods on Original and JPEG Compressed Carvalho Images. For each
column, we highlight in bold the top-performing method.}}

\noindent \begin{raggedright}
\hspace{-0.75in}%
\begin{tabular}{|c|c|c|c|}
\hline 
Method & Original (No Compression) & JPEG Quality = 70 & JPEG Quality = 50\tabularnewline
\hline 
\hline 
SFCN & 0.4411 & 0.4350 & 0.4326\tabularnewline
\hline 
MFCN & 0.4678 & 0.4434 & 0.4334\tabularnewline
\hline 
Edge-enhanced MFCN & \textbf{0.4795} & \textbf{0.4496} & \textbf{0.4431}\tabularnewline
\hline 
NOI1 & 0.3430 & 0.3284 & 0.3292\tabularnewline
\hline 
DCT & 0.3066 & 0.3103 & 0.3121\tabularnewline
\hline 
CFA2 & 0.3124 & 0.2850 & 0.2832\tabularnewline
\hline 
NOI3 & 0.2693 & 0.2646 & 0.2636\tabularnewline
\hline 
BLK & 0.3069 & 0.2946 & 0.3005\tabularnewline
\hline 
ELA & 0.2756 & 0.2703 & 0.2728\tabularnewline
\hline 
ADQ1 & 0.2943 & 0.2677 & 0.2646\tabularnewline
\hline 
CFA1 & 0.2932 & 0.2901 & 0.2919\tabularnewline
\hline 
NOI2 & 0.3155 & 0.2930 & 0.2854\tabularnewline
\hline 
\end{tabular}
\par\end{raggedright}
\centering{}\label{jpeg_compress}
\end{table}

\subsection{Performance on Gaussian Blurred Images}

We also evaluated the performance of the proposed and existing methods
after applying Gaussian blurring or smoothing to the spliced images
of the Carvalho dataset. We filtered a given spliced image using a
Gaussian smoothing kernel with the following four standard deviation
values (in terms of pixels): $\sigma=0.5$, $1.0$, $1.5$, and $2.0$.
Table \ref{blurring} shows the average $F_{1}$ scores of the proposed
methods and existing methods applied to the original and Gaussian-filtered
images. We see slight performance degradation when $\sigma=2.0.$
However, the performance of the SFCN and MFCN methods is still better
than the performance of existing methods. Furthermore, we see that
the performance of the SFCN and MFCN methods on the blurred images
is better than the performance of existing methods on the original
unblurred images. 
\begin{table}
\caption{\textcolor{black}{Average $F_{1}$ Scores of Proposed and Existing
Methods On Original and Blurred Carvalho Images. For each column,
we highlight in bold the top-performing method.}}

\noindent \begin{raggedright}
\hspace{-0.5in}%
\begin{tabular}{|c|c|c|c|c|c|}
\hline 
Method & Original (No Blurring) & $\sigma=0.5$ & $\sigma=1.0$ & $\sigma=1.5$ & $\sigma=2.0$\tabularnewline
\hline 
\hline 
SFCN & 0.4411 & 0.4403 & 0.4475 & 0.4365 & 0.4219\tabularnewline
\hline 
MFCN & 0.4678 & 0.4694 & 0.4659 & 0.4560 & 0.4376\tabularnewline
\hline 
Edge-enhanced MFCN & \textbf{0.4795} & \textbf{0.4849} & \textbf{0.4798} & \textbf{0.4724} & \textbf{0.4482}\tabularnewline
\hline 
NOI1 & 0.3430 & 0.3330 & 0.2978 & 0.2966 & 0.2984\tabularnewline
\hline 
DCT & 0.3066 & 0.3040 & 0.3014 & 0.3013 & 0.2994\tabularnewline
\hline 
CFA2 & 0.3124 & 0.3055 & 0.2947 & 0.2907 & 0.2894\tabularnewline
\hline 
NOI3 & 0.2693 & 0.2629 & 0.2539 & 0.2500 & 0.2486\tabularnewline
\hline 
BLK & 0.3069 & 0.3133 & 0.3177 & 0.3177 & 0.3168\tabularnewline
\hline 
ELA & 0.2756 & 0.2740 & 0.2715 & 0.2688 & 0.2646\tabularnewline
\hline 
ADQ1 & 0.2943 & 0.2929 & 0.2922 & 0.2952 & 0.2960\tabularnewline
\hline 
CFA1 & 0.2932 & 0.2974 & 0.3085 & 0.3022 & 0.3056\tabularnewline
\hline 
NOI2 & 0.3155 & 0.3141 & 0.3074 & 0.3040 & 0.2982\tabularnewline
\hline 
\end{tabular}
\par\end{raggedright}
\centering{}\label{blurring}
\end{table}

\subsection{Performance on Images with Additive White Gaussian Noise}

Finally, we evaluated the performance of the proposed and existing
methods on images with additive white gaussian noise (AWGN). We added
AWGN to the images in the Carvalho testing set and set the resulting
SNR values to three levels: 25 dB, 20 dB, and 15 dB. Table \ref{noise}
shows the $F_{1}$ scores of the proposed methods and existing methods
on the original and the noisy images. Again, we see small degradation
in the performance of the proposed methods due to additive noise.
However, the performance of the proposed SFCN and MFCN methods is
still better than the performance of existing methods. Furthermore,
we see that the performance of the SFCN and MFCN methods on the corrupted
images is better than the performance of existing methods on the original
uncorrupted images. 
\begin{table}
\caption{\textcolor{black}{Average $F_{1}$ Scores of Proposed and Existing
Methods On Original and Noisy Carvalho Images. For each column, we
highlight in bold the top-performing method.}}

\noindent \begin{raggedright}
\hspace{-0.55in}%
\begin{tabular}{|c|c|c|c|c|}
\hline 
Method & Original (No Noise) & SNR = 25 dB & SNR = 20 dB & SNR = 15 dB\tabularnewline
\hline 
\hline 
SFCN & 0.4411 & 0.4400 & 0.4328 & 0.4307\tabularnewline
\hline 
MFCN & 0.4678 & 0.4674 & 0.4677 & 0.4577\tabularnewline
\hline 
Edge-enhanced MFCN & \textbf{0.4795} & \textbf{0.4786} & \textbf{0.4811} & \textbf{0.4719}\tabularnewline
\hline 
NOI1 & 0.3430 & 0.3181 & 0.3038 & 0.2918\tabularnewline
\hline 
DCT & 0.3066 & 0.2940 & 0.2681 & 0.2485\tabularnewline
\hline 
CFA2 & 0.3124 & 0.2863 & 0.2844 & 0.2805\tabularnewline
\hline 
NOI3 & 0.2693 & 0.2535 & 0.2496 & 0.2476\tabularnewline
\hline 
BLK & 0.3069 & 0.2974 & 0.2813 & 0.2658\tabularnewline
\hline 
ELA & 0.2756 & 0.2533 & 0.2473 & 0.2460\tabularnewline
\hline 
ADQ1 & 0.2943 & 0.2916 & 0.2874 & 0.2937\tabularnewline
\hline 
CFA1 & 0.2932 & 0.2909 & 0.2906 & 0.2859\tabularnewline
\hline 
NOI2 & 0.3155 & 0.3207 & 0.3163 & 0.3108\tabularnewline
\hline 
\end{tabular}
\par\end{raggedright}
\centering{}\label{noise}
\end{table}

\section{Conclusion}

\label{sec:conclusion}

It was demonstrated in this work that the application of FCN to the
splicing localization problem yields large improvement over current
published techniques. The FCN we utilized is based on the FCN VGG-16
architecture with skip connections, and we incorporated several modifications,
such as batch normalization layers and class weighting. We first evaluated
a single-task FCN (SFCN) trained only on the surface ground truth
mask (which classifies each pixel in an image as spliced or authentic).
Although the single-task network is shown to outperform existing techniques,
it can still yield a coarse localization output in certain cases.
Thus, we next proposed the use of a multi-task FCN (MFCN) that is
simultaneously trained on the surface ground truth mask and the edge
ground truth mask, which indicates whether each pixel belongs to the
boundary of the spliced region. For the MFCN-based method, we presented
two different inference approaches. In the first approach, we compute
the binary system output mask by thresholding the surface output probability
map. In this approach, the edge output probability map is not utilized
in the inference step. This first MFCN-based inference approach is
shown to outperform the SFCN-based approach. In the second MFCN-based
inference approach, which we refer to as edge-enhanced MFCN, we utilize
both the surface and edge output probability map when generating the
binary system output mask. The edge-enhanced MFCN is shown to yield
finer localization of the spliced region, as compared to the SFCN-based
approach and the MFCN without edge-enhanced inference. The proposed
methods were evaluated on manipulated images from the Carvalho, CASIA
v1.0, Columbia, and the DARPA/NIST Nimble Challenge 2016 SCI datasets.
The experimental results showed that the proposed methods outperform
existing splicing localization methods on these datasets, with the
edge-enhanced MFCN performing the best. 

\section{Acknowledgments}

This material is based on research sponsored by DARPA and Air Force
Research Laboratory (AFRL) under agreement number FA8750-16-2-0173.
The U.S. Government is authorized to reproduce and distribute reprints
for Governmental purposes notwithstanding any copyright notation thereon.
The views and conclusions contained herein are those of the authors
and should not be interpreted as necessarily representing the official
policies or endorsements, either expressed or implied, of DARPA and
Air Force Research Laboratory (AFRL) or the U.S. Government.

Credits for the use of the CASIA Image Tempering Detection Evaluation
Database (CASIA TIDE) v1.0 and v2.0 are given to the National Laboratory
of Pattern Recognition, Institute of Automation, Chinese Academy of
Science, Corel Image Database and the photographers. http://forensics.idealtest.org

\bibliographystyle{elsarticle-harv}
\bibliography{references}

\end{document}